\pdfoutput=1

\documentclass{article}

\usepackage[nonatbib]{nips_2016}

\usepackage[%
    minbibnames=1, maxbibnames=99,
    style=alphabetic,
    doi=false,url=false,
    firstinits=true,hyperref,natbib,backend=bibtex]{biblatex}
\renewbibmacro{in:}{%
  \ifentrytype{article}{}{\printtext{\bibstring{in}\intitlepunct}}}
\bibliography{biblio}

\usepackage[labelfont=bf]{caption}

\usepackage[usenames,dvipsnames]{xcolor}
\usepackage[utf8]{inputenc} 
\usepackage{amsmath, amssymb, bm, mathtools, thmtools}
\usepackage{verbatim}
\usepackage{graphicx}\graphicspath{{figures/}}
\usepackage{multicol}
\usepackage{tabularx}
\usepackage{mathrsfs} 
\usepackage{url}

\usepackage[colorlinks,citecolor=blue,urlcolor=blue,linkcolor=blue]{hyperref}
\usepackage{hypernat}
\usepackage{datetime}

\DeclareMathAlphabet\EuRoman{U}{eur}{m}{n}
\SetMathAlphabet\EuRoman{bold}{U}{eur}{b}{n}

\usepackage{euscript,microtype}
\usepackage[capitalize]{cleveref}

\crefname{lemma}{Lemma}{Lemmas}
\crefname{corollary}{Corollary}{Corollaries}
\crefname{theorem}{Theorem}{Theorems}

\makeatletter
\let\reftagform@=\tagform@
\def\tagform@#1{\maketag@@@{\ignorespaces\textcolor{gray}{(#1)}\unskip\@@italiccorr}}
\renewcommand{\eqref}[1]{\textup{\reftagform@{\ref{#1}}}}
\makeatother

\newcommand{\LATER}[1]{\error}
\newcommand{\fLATER}[1]{\error}
\newcommand{\TBD}[1]{\error}
\newcommand{\fTBD}[1]{}
\newcommand{\PROBLEM}[1]{\error}
\newcommand{\fPROBLEM}[1]{}

\def\[#1\]{\begin{align}#1\end{align}}
\def\*[#1\]{\begin{align*}#1\end{align*}}

\DeclareMathOperator*{\newlim}{\mathrm{lim}\vphantom{\mathrm{infsup}}}

\DeclareMathOperator*{\newmax}{\mathrm{max}\vphantom{\mathrm{infsup}}}

\renewcommand{\lim}{\newlim}

\renewcommand{\max}{\newmax}

\definecolor{linenocolor}{rgb}{.85,0.85,.85}

\newcommand{\prob}[2]{p_w(#1|#2)}
\newcommand{\toplabel}[1]{\ell_{#1}}
\newcommand{\JPG}[1]{\mathrm{JPG}(#1)}
\newcommand{\ADV}[1]{\mathrm{Adv}_{\epsilon}(#1)}
\newcommand{\ADVE}[2]{\mathrm{Adv}_{#2}(#1)}
\newcommand{\JPGnoise}[1]{\mathrm{JPG_{noise}}(#1)}

\newcommand{\jpgnoise}[1]{\mathrm{\eta_{\text{JPG}}}(#1)}

\title{A study of the effect of JPG compression on adversarial images}

\author{Gintare Karolina Dziugaite \\
Department of Engineering \\
University of Cambridge\\
\And
Zoubin Ghahramani \\
Department of Engineering \\
University of Cambridge\\
\And
Daniel M. Roy \\
Department of Statistical Sciences \\
University of Toronto\\
}

\newcommand{\MLprob}{top-label probability}
\newcommand{\MLprobs}{top-label probabilities}

\begin{document}

\maketitle

\begin{abstract}
Neural network image classifiers are known to be vulnerable to adversarial images, i.e., 
natural images which have been modified by an adversarial perturbation specifically designed to be imperceptible to humans
yet fool the classifier.
Not only can adversarial images be generated easily, but these images will often be adversarial for networks trained on disjoint subsets of data or with different architectures.
Adversarial images represent a potential security risk as well as a serious machine learning challenge---it is clear that vulnerable neural networks perceive images very differently from humans.
Noting that virtually every image classification data set is composed of JPG images,
we evaluate the effect of JPG compression on the classification of adversarial images.
For Fast-Gradient-Sign perturbations of small magnitude, we found that JPG compression often reverses the drop in classification accuracy to a large extent, but not always.  As the magnitude of the perturbations increases, JPG recompression alone is insufficient to reverse the effect.
\end{abstract}

\section{Introduction}

Neural networks are now widely used across machine learning, 
including image classification, where they achieve state-of-the-art accuracy on standard benchmarks \citep{ImageNet,HeZRS15}. 
However, neural networks have recently been shown to be vulnerable to \emph{adversarial} examples \citep{Szegedy2014},
i.e., inputs to the network that have undergone imperceptible perturbations specifically optimized to cause the neural network to strongly misclassify. 

Most neural networks trained for image classification are trained on images that have undergone JPG compression. 
Adversarial perturbations are unlikely to leave an image in the space of JPG images, and so this paper explores the idea that JPG (re)compression 
could remove some aspects of the adversarial perturbation.
Our experiments show that JPG compression often succeeds in reversing the adversarial nature of images that have been modified by a small-magnitude perturbation produced by the Fast Gradient Sign method of \citet{Goodfellow14}.
However, as the magnitude of the perturbation increases, JPG compression is unable to recover a non-adversarial image and therefore JPG compression cannot, by itself, guard against the security risk of adversarial examples.

We begin by discussing related work and in particular a recent preprint by \citet{Kurakin2016} showing independent work that the effect of certain varieties of adversarial perturbations can even survive being printed on paper and recaptured by a digital camera.  This same preprint also reports on the effect of JPG compression quality on adversarial perturbations.  Our experiments are complimentary, as we vary the magnitude of the perturbation.

\section{Related Work}

\citet{Szegedy2014} were the first to demonstrate adversarial examples: working within the context of image classification,  
they found the smallest additive perturbation $\eta$ to an image $x$ that caused the network to misclassify the image $x + \eta$.
In their paper introducing the concept, they demonstrated the surprising phenomenon that adversarial examples generalized across neural networks trained on disjoint subsets of training data, as well as across neural networks with different architectures and initializations. 
\citet{PapernotMGJCS16} exploited this property to demonstrate how one could construct adversarial examples for a network of an unknown architecture by training an auxiliary neural network on related data. 

These findings highlight that adversarial examples pose a potential security risk in real-world applications of neural networks such as autonomous car navigation and medical image analysis.
Adversarial examples also pose a challenge for machine learning, because they expose an apparently large gap between the inductive bias of humans and machines.
In part due to both challenges, there has been a flood of work following the original demonstration of adversarial examples that attempts to explain the phenomenon and protect systems.

\citet{Goodfellow14} argued that neural networks are vulnerable to adversarial perturbations 
due to the linear nature of neural networks and presented some experimental evidence that neural network classifiers with non-linear activations are more robust. 
\citet{Tabacof15} demonstrated empirically that adversarial examples are not isolated points and the neural networks are more robust to random noise than adversarial noise. \citet{Billovits2016} visualized how adversarial perturbations change activations in a convolutional neural network. They also ran a number of experiments to better understand which images are more susceptible to adversarial perturbations depending on the magnitude of the classifier's prediction on  clean versions of the image.

Several authors have proposed solutions to adversarial examples with mixed success \citep{PapernotMWJS15, GuR14}. \citet{GuR14} proposed the use of an autoencoder (AE) to remove adversarial perturbations from inputs. 
 While the AE could effectively remove adversarial noise, the combination of the AE and the neural network was even less robust to adversarial perturbations. 
 They proposed to use a contractive AE instead, which increased the size of the perturbation needed to alter the classifier's predicted class.

While most of the work has been empirical, \citet{Fawzi15} gave a theoretical analysis of robustness to adversarial examples and random perturbations for binary linear and quadratic classifiers. They compute upper bounds on the robustness of linear and quadratic classifiers. The upper bounds suggests that quadratic classifiers are more robust to adversarial perturbations than linear ones.

A recent paper by \citet{Kurakin2016} makes several significant contributions to the understanding of adversarial images.
In addition to introducing several new methods for producing large adversarial perturbations that remain imperceptible,
they demonstrate the existence of adversarial examples ``in the physical world''.
To do so, \citeauthor{Kurakin2016} compute adversarial images for the Inception classifier \citep{SzegedyVISW15},
print these adversarial images onto \emph{paper},
and then recapture the images using a cell-phone camera.
They demonstrate that, even after this process of printing and recapturing, 
a large fraction of the images remain adversarial. 
The authors also experimented with multiple transformations of adversarial images: changing brightness and contrast,  
adding Gaussian blur, and varying JPG compression quality.  This last aspect of their work relates to the experiments we report here.

\section{Hypothesis}

\begin{figure}[t] 
\centering
\includegraphics[width=.7\linewidth]{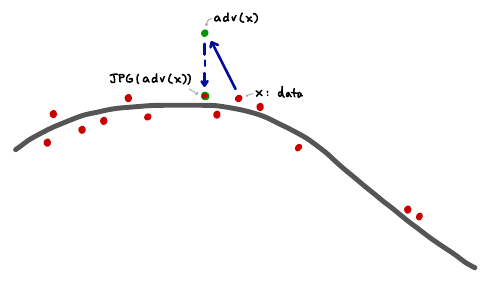}
\caption{
The red dots represent the data and the grey line the data subspace. The solid blue arrow is the adversarial perturbation that moves the data point $x$ away from the data subspace and the dotted blue arrow is the projection on the subspace. In the case where the perturbation is approximately orthogonal to the JPG subspace, JPG compression brings the adversarial example back to the data subspace.
}
\label{jpgprojection}
\end{figure}

What is the nature of adversarial examples? Why do they exist?  And why are they  robust to changes in training data, network architecture, etc?

Adversarial perturbations are considered interesting because they are judged to be imperceptible by humans, yet they are (by definition) extremely perceptible to neural network classifiers, even across a wide variety of training regimes.
A basic hypothesis underlying this work is that, 
in any challenging high-dimensional classification task where the inputs naturally live in (or near) a complex lower-dimensional \emph{data subspace}, 
adversarial examples will lie outside this data subspace, taking advantage of the fact that the training objective for the neural network is essentially agnostic to the network's behavior outside the data subspace.

Even if individual neural network classifiers were not robust to imperceptible perturbations, 
we might settle for a measure of confidence/credibility reporting high uncertainty on adversarial examples.
In theory, we would expect confidence intervals or credible sets associated with neural network classifiers to represent high uncertainty on adversarial images \emph{provided} that, outside the data subspace, there was disagreement among the family of classifiers achieving, e.g., high likelihood/posterior probability. In practice, efficient computational methods may not be able to determine whether there is uncertainty.  The field has poor understanding of both issues.
To date, no frequentist or Bayesian approach has demonstrated the ability to correctly classify or report high uncertainty on adversarial images.

At the very least, adversarial examples reflect the fact that neural network classifiers are relying on properties of the data different from those used by humans.
In theory, even a classifier trained on a data set of diverging size might fall prey to adversarial examples if the training data live on a subspace. Techniques such as data augmentation (e.g., by adding noise or adversarial perturbations) would be expected to remove a certain class of adversarial examples, but unless the notion of ``perceptible perturbation'' is exactly captured by the data augmentation scheme, it seems that there will always be space for adversarial examples to exist.\footnote{The extent to which humans are themselves susceptible  to adversarial imagery is not well understood, at least by the machine learning community.  Can small perturbations (e.g., in the mean-squared-error) cause human perception to change dramatically?}

Natural image classification is an example of a high-dimensional classification task whose inputs have low intrinsic dimension.  Indeed, we can be all but certain that if we were to randomly generate a bitmap, the result would not be a natural image. On the other hand, humans are not affected by adversarial perturbations or other perturbations such as random noise, and so we introduce the notion of the  \emph{perceptual subspace}: the space of bitmaps perceived by humans as being natural images with some corruption.
Empirical evidence suggests that neural networks learn to make accurate predictions inside the data subspace.  Neural networks are also understood to be fairly resistant to random perturbations as these perturbations are understood to cancel themselves out \citep{Goodfellow14}. 
Neural networks classifiers work well, in part, due to their strong inductive biases.  But this same bias means that a neural network may report strong predictions beyond the data subspace where there is no training data. We cannot expect sensible predictions outside the data subspace from individual classifiers.\footnote{
One would hope that, even if individual neural networks achieving high posterior probability suffered from adversarial perturbations, 
networks sampled from a Bayesian posterior would disagree on the classification of an input outside the data subspace, representing uncertainty.  However, our experiments with current scalable approximate Bayesian neural network methods (namely, variants of stochastic gradient Langevin dynamics \citep{WelTeh2011a,pSGLD}) revealed that Bayesian neural networks report confident misclassifications on adversarial examples.  It is worth evaluating other approximate inference frameworks.}

If we could project adversarial images back onto the data subspace, we could conceivable get rid of adversarial perturbations.
Unfortunately, it is not clear whether it is possible to characterize or learn a suitable representation of the data subspace corresponding to natural images.
We may, however, be able to find other lower-dimensional subspaces that contain the data subspace.
To that end, note that most image classification data sets, like ImageNet \citep{ImageNet}, are built from JPG images. Call this set of images the \emph{JPG subspace}, which necessarily contains the data subspace.
Perturbations of natural images (by adding scaled white noise or randomly corrupting a small number of pixels) are almost certain to move an image out of the JPG subspace and therefore out of the data subspace.  
While we cannot project on the data subspace, we can use JPG compression to ``project'' the perturbed images 
back onto the JPG subspace.  
We might expect JPG compression to reverse adversarial perturbations for several reasons: 
First, adversarial perturbations could be very sensitive and reversed by most image processing steps. (Our findings contradict this, as do the findings in \citep{Kurakin2016}.)
Second, adversarial perturbations might be ``orthogonal'' to the JPG subspace, in which case we would expect the modifications to be removed by JPG compression. (Our findings for small perturbations do not contradict this idea, though larger perturbations are not removed by JPG compression.  It would be interesting to evaluate the discrete cosine transformation of adversarial images to settle this hypothesis.)
More study is necessary to explain our findings. 

\section{Empirical findings}

We evaluated the effect of adversarial perturbations on the network's classification, and then studied how the classification was affected by a further JPG compression of the adversarial image.  We measured the change at several different magnitudes of adversarial perturbation. 

We used the pre-trained OverFeat network (Sermanet et al., 2013), which was trained on images from the  2012 ImageNet training set (1000 classes).
The training images used to produce OverFeat underwent several preprocessing steps:
they were scaled so that the smallest dimension was 256;
then 5 random crops of size $221 \times 221$ were produced;
finally, the set of images (viewed as vectors) were then standardized to have zero mean and unit variance.  (When we refer to standardization below, we are referring to the process of repeating precisely the same shift and scaling used to standardize the training data fed to OverFeat.)
The OverFeat network is composed of ReLU activations and max pooling operations, 5 convolutional layers, and 3 fully connected layers. 

For a (bitmap) image $x$, we will write $\JPG{x}$ to denote the JPG compression of $x$ at quality level $75$.
For a network with weights $w$ and input image $x$, let $\prob{c}{x}$ be the probability assigned to class $c$. 
Let $\toplabel{x} = \arg\max \prob{c}{x}$ be the class label assigned the highest probability (which we will assume is unique).
Then $\prob{\toplabel{x}}{x}$ is the probability assigned to this label. 

\begin{figure}[t] 
\centering
\includegraphics[width=.194\linewidth]{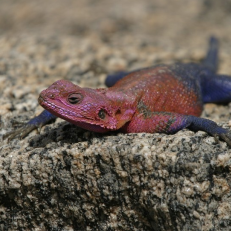}
\includegraphics[width=.194\linewidth]{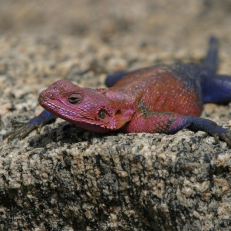}
\includegraphics[width=.194\linewidth]{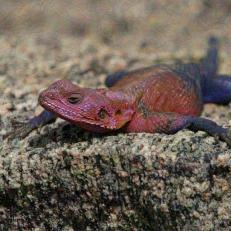}
\includegraphics[width=.194\linewidth]{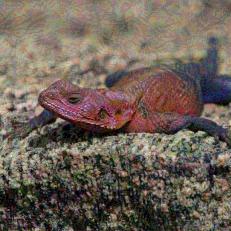}
\includegraphics[width=.194\linewidth]{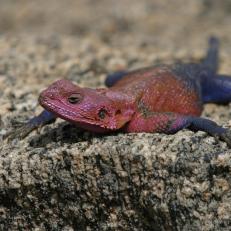}
\caption{
(\textit{first}) Original image $x$, with label ``agama'' assigned 0.99 probability; 
(\textit{second}) Adversarial image $\ADV{x}$, where $\epsilon = 1$, with label ``rock crab'' assigned 0.93 probability and label ``agama'' assigned $6 \times 10^{-5}$ probability; 
(\textit{third and fourth}) Adversarial images $\ADV{x}$ with $\epsilon$ set to 5 and 10. Both assign probability $\approx 0$ to ``agama''. However, adversarial noise becomes apparent;
(\textit{last}) JPG compression of the adversarial image, $\JPG{\ADV{x}}$ with $\epsilon =1$, with label ``agama'' assigned 0.96 probability.
}
\label{exampleimg}
\end{figure}

To generate adversarial examples, we used the Fast Gradient Sign method introduced by \citet{Goodfellow14}. 
Let $w$ represent the pre-trained weights of the OverFeat network.
The Fast Gradient Sign perturbation is calculated by scaling the element-wise sign of the gradient of the training objective $J(x, w, y)$ with respect to the image $x$ for the label $y=\toplabel{x}$, i.e.,
$$
\eta_\epsilon (x)
=  \frac{\epsilon}{255} \, \text{sign} \bigl ( \nabla_{x'} J(x', w, y) \mid_{x'=x,y=\toplabel{x}}  \bigr )
$$
and thus
\[
\ADV{x} 
= x + \eta_\epsilon(x)
\]
The image gradient $\nabla_{x'} J(x', w, y)$ can be efficiently computed using back propagation. In our experiments with the OverFeat network, we used $\epsilon \in \{1,5,10 \}$.
See \cref{exampleimg} for several examples of images after adversarial perturbations of increasing magnitudes.

\newpage
For each image $x$ in the ImageNet validation set, we performed the following steps:
\begin{enumerate}
	\item Scale $x$ so that its smallest dimension is $256$; crop to the centered $221 \times 221$ square region; and then standardize;
	\item Compute $\ADV{x}$ using the Fast Gradient Sign method, with $\epsilon \in \{1,5,10\}$;
	\item Compute $\JPG{\ADV{x}}$ using the \texttt{save} method from Torch7's \texttt{image} package;
	\item Compute the OverFeat network predictions for all images: original $x$, adversarial $\ADV{x}$; and compressed $\JPG{\ADV{x}}$.
\end{enumerate}

For an image $x$, we will refer to $\prob{\toplabel{x}}{x}$ as its \emph{\MLprob} and, more generally, for a transformation $f$ acting on images,
we will refer to $\prob{\toplabel{x}}{f(x)}$ as the \emph{\MLprob\ after transformation $f$}.

\begin{figure}[t]
\centering
\includegraphics[width=\linewidth]{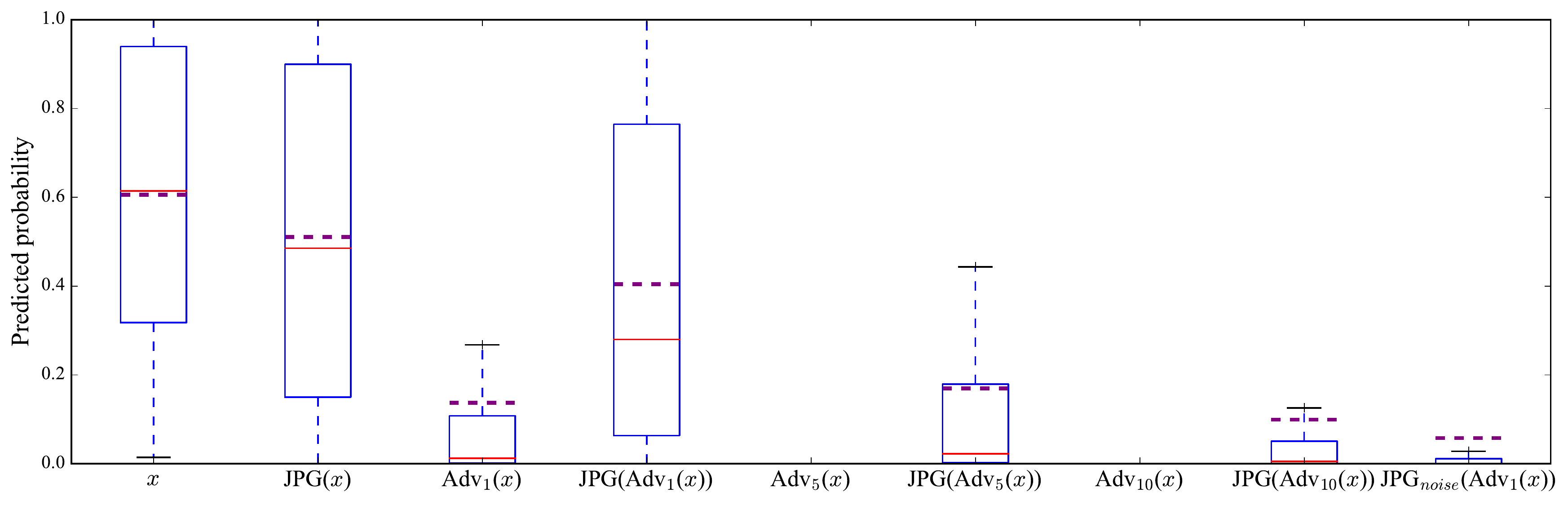}
\caption{The \MLprobs, i.e., the \emph{predicted probability} (y-axis) assigned to the most likely class $\toplabel{x}$, after various transformations $x \mapsto f(x)$. The red horizontal line in each box plots is the average \MLprob. The solid red line is the median, the box represents the interquartile range, and the whiskers represent the minimum and maximum values, excluding outliers. Labels along the bottom specify the transformation $f(x)$ applied to the image $x$ before measuring the \MLprob.}
\label{boxplots}
\end{figure}

\begin{figure}[t]
\centering
\includegraphics[width=\linewidth]{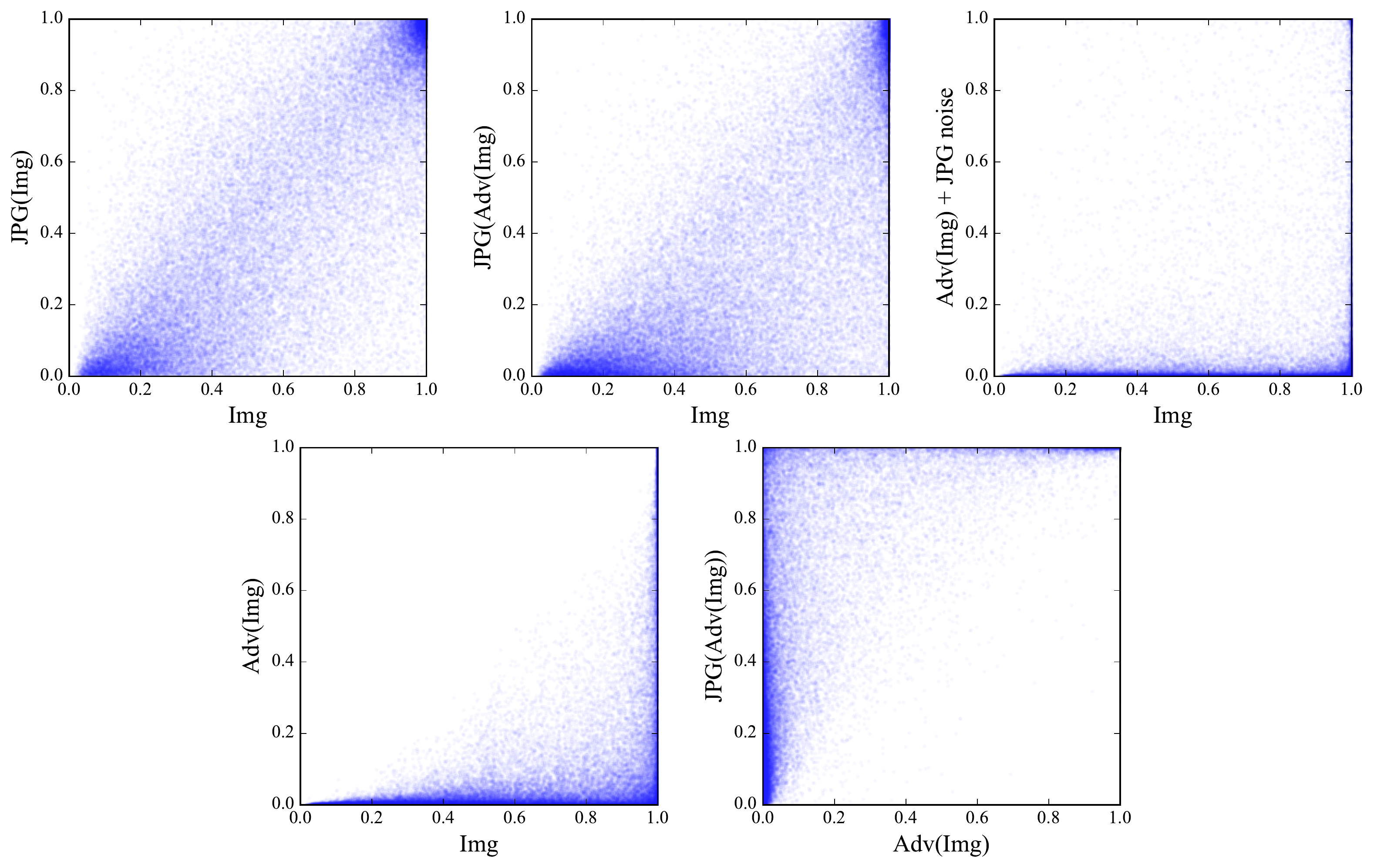}
\caption{
In every scatter plot, every validation image $x$ is represented by a point $(p_1,p_2)$, which specifies the \MLprobs\ $p_j = \prob{\toplabel{x}}{f_j(x)}$ under a pair $(f_1,f_2)$ of modifications of the image, respectively. 
All adversarial perturbations in these figures were generated with magnitude $\epsilon =1$.
Along the top row, the $x$-axis represents the \MLprob\ for a clean image. 
(\textit{top left}) The plot illustrates the effect of JPG compression of a natural image. The predictions do change, but on average they lie close to the diagonal and do not change the \MLprob\ appreciably;
(\textit{top middle}) If JPG compression of the adversarial image removed adversarial perturbations, we would expect this plot to look like the one to the left. While they are similar (most points lie around the diagonal), more images lie in the lower right triangle, suggesting that the adversarial perturbations are sometimes not removed or only partially removed.
(\textit{top right}) Adding JPG noise does not reverse the effect of adversarial perturbations: indeed, points lie closer to the lower axis than under a simple adversarial modification;
(\textit{bottom left}) The \MLprobs\ after adversarial perturbation drops substantially on average; 
(\textit{bottom right}) This plot complements the top-middle plot.  Most of the points lie on the upper left triangle, which suggests that JPG compression of an adversarial image increases the \MLprob\ and partially reverses the effect of many adversarial perturbations. 
}
\label{scatterplot}
\end{figure}

\cref{boxplots} gives a coarse summary of how JPG compression affects adversarial examples,
while \cref{scatterplot} gives a more detailed picture at the level of individual images for the case of perturbations of magnitude $\epsilon=1$.  We will now explain these figures in turn.

\cref{boxplots} reports statistics on the \MLprob\ under various transformations for every image in the validation set .
The first boxplot summarizes the distribution of the \MLprob\ for the validation images when no perturbations have been made.  As we see, the network assigns, on average, 0.6 probability to the most probable label and the interquartile range lies away from the extremes $0$ and $1$.
While we might consider JPG (re)compression to be a relatively innocuous operation, the second boxplot reveals that JPG compression already affects the \MLprob\ negatively.
The third boxplot summarizes the \MLprob\ under an adversarial transformation of magnitude $1/255$: the mean probability assigned to the top label $\toplabel{x}$ drops from approximately $0.6$ to below $0.15$.
The \MLprob\ after JPG compression of the adversarial images increases back towards the levels of JPG compressed images, but falls short: the mean recovers to just over 0.4.
Larger adversarial perturbations (of magnitude $5/255$ and $10/255$) cause more dramatic negative changes to the \MLprob.  Moreover, JPG compression of these more perturbed images is not effective at reversing the adversarial perturbation: the \MLprob\ remains almost unchanged, improving only slightly. 

The scatter plots in \cref{scatterplot} paint a more detailed picture for small advesarial perturbations ($\epsilon=1$).
In every scatter plot, a point $(p_1,p_2)$ specifies the \MLprob\ under a pair $(f_1,f_2)$ of transformations, respectively. 
In the first plot, we see the effect of JPG compression on the \MLprob, which can be combined with the second boxplot in \cref{boxplots} to better understand the effect of JPG compression on a neural networks \MLprob\ assignments.  In short, JPG compression can lower and raise the \MLprob, although the mean effect is negative, and JPG compression affects images with high \MLprobs\ least.
The bottom-left plot shows the strong negative effect of the adversarial perturbation on the \MLprob, 
which can be contrasted with the top-middle plot, where we see that the \MLprobs\ recover almost to the level of the original images after JPG recompression. (C.f., boxplots 2 and 4 in \cref{boxplots}.)

If JPG compression were a good surrogate for projection onto the data subspace, we would expect
the \MLprobs\ to recover to the level of the \MLprobs\ for $\JPG{x}$.
This is not quite the case, even for small perturbations ($\epsilon = 1$), although the adversarial nature of these images is often significantly reduced.  For larger perturbations, the effect of JPG compression is small.  (This agrees with the finding by \citet{Kurakin2016} that Fast Gradient Sign perturbations are quite resilient to image transformations, including JPG compression.)

Does the improvement for small perturbations yielded by JPG compression 
depend on the specific structure of JPG compression or could it be mimicked with noise sharing some similar statistics?
To test this hypothesis, we studied the effect on \MLprobs\ after adding a random permutation of the vector representing the effect of JPG compression.
More precisely, let P be a random permutation matrix.
We tested the effect of the perturbation
\[
\jpgnoise{x} =  P \Delta (\ADV{x}), \qquad \text{where $ \Delta (x')  = \JPG{x'} - x' $} ,
\]
which we call \emph{JPG noise}. Thus, we studied the \MLprobs\ for images of the form
\[
\JPGnoise{\ADV{x}} = \ADV{x} + \jpgnoise{x}.
\]
By construction, JPG noise
 shares every permutation-invariant statistics with JPG compression, but loses, e.g., information about the direction of the JPG compression modification. 
 The last box plot in 
\cref{boxplots} 
shows that adversarial images remain adversarial after adding JPG noise: indeed, the average predicted probability for $\toplabel{x}$ is even lower than for adversarial images (second box plot). 

\cref{resultstable} summarizes classification accuracy and mean \MLprobs\ after various transformations applied to images in the ImageNet validation set. (C.f., \cref{boxplots}.) Notice that the accuracy drops dramatically after adversarial perturbation. JPG compression increases the accuracy substantially for small perturbations ($\epsilon = 1$), however, the accuracy is still lower than on clean images. 
For larger adversarial perturbations ($\epsilon \in \{5,10\}$), JPG compression does not increase accuracy enough to represent a practical solution to adversarial examples.

\begin{table}[t]
\centering
\begin{tabular}{rcc}
             Modification $f(x)$ &  Top-1 Accuracy   &  Mean $\prob{\toplabel{x}}{f(x)}$  \\
        & & \\
 $x$\hspace*{.75em}   & 0.58 & 0.61   \\        
$\ADVE{x}{1}$\hspace*{.4em}       & 0.23 & 0.13       \\ 
$\ADVE{x}{5}$\hspace*{.4em}       & 0.11 & 0.04       \\ 
$\ADVE{x}{10}$\hspace*{.4em}      & 0.09 & 0.04       \\ 
$\JPG{\ADVE{x}{1}}$  & 0.48 & 0.41             \\ 
$\JPG{\ADVE{x}{5}}$  & 0.26 & 0.17             \\ 
$\JPG{\ADVE{x}{10}}$  & 0.17 & 0.04             \\ 
$\JPGnoise{\ADVE{x}{1}}$  & 0.07 & 0.06        \\ 
\\
\end{tabular}
\caption{
Classification accuracy and mean \MLprobs\ after various transformations.
}
\label{resultstable}
\end{table}

\section{Conclusion}

Our experiments demonstrate that JPG compression can reverse small adversarial perturbations created by the Fast-Gradient-Sign method. 
However, if the adversarial perturbations are larger, JPG compression does not reverse the adversarial perturbation.
In this case, the strong inductive bias of neural network classifiers leads to incorrect yet confident misclassifications.
Even the largest perturbations that we evaluated are barely visible to an untrained human eye, and so JPG compression is far from a solution.
We do not yet understand why JPG compression reverses small adversarial perturbations.

\section*{Acknowledgments}

ZG acknowledges funding from the Alan Turing Institute, Google, Microsoft Research and EPSRC Grant EP/N014162/1.
DMR is supported in part by a Newton Alumni grant through the Royal Society.

\printbibliography

\vfill
\end{document}